\documentclass[letterpaper, 10 pt, conference]{ieeeconf}  

\IEEEoverridecommandlockouts                              

\usepackage[usenames,dvipsnames,table,xcdraw]{xcolor}  
\usepackage{hyperref}
\usepackage{amsmath,amsfonts}
\usepackage{algorithm}                    
\usepackage{algorithmicx}                  
\usepackage[noend]{algpseudocode}
\usepackage{graphicx}
\usepackage{multirow}
\usepackage{subcaption}
\usepackage{svg}
\usepackage{array,booktabs,makecell,nicematrix,graphicx,amssymb,xcolor}
\usepackage{tikz}
\usepackage[normalem]{ulem}
\useunder{\uline}{\ul}{}

\newcommand{\cspace}{\ensuremath{\mathcal{C}_{space}}}
\newcommand{\cspaces}{\ensuremath{\mathcal{C}_{spaces}}}

\newcommand{\compcspace}{\ensuremath{\mathcal{C}_{comp}}}

\newcommand{\simd}{{\sc SIMD}}

\newcommand{\mrmp}{{\sc MRMP}}

\newcommand{\arc}{{\sc ARC}}
\newcommand{\parc}{{\sc P-ARC}}
\newcommand{\orarc}{{\sc OR-ARC}}
\newcommand{\orparc}{{\sc OR-P-ARC}}

\title{\LARGE \bf
P-ARC: Exploiting Subproblem Independence for Parallel Multi-Robot Motion Planning
}

\author{
James D. Motes$^{1}$,  Marco Morales$^{1,2}$, and Nancy M. Amato$^{1}$
\thanks{$^{1}$James D. Motes, Marco Morales, and Nancy M. Amato are with the Parasol Lab, School of Computing and Data Science, University of Illinois at Urbana Champaign, Champaign, IL, 61820 USA.
{\tt\small jmotes2, moralesa, namato@illinois.edu}}%
\thanks{$^{2}$Marco Morales is also with the Department of Computer Science at Instituto Tecnol\'ogico Aut\'onomo de M\'exico (ITAM), Mexico City, M\'exico.}
}

\begin{document}

\maketitle
\thispagestyle{empty}
\pagestyle{empty}

\begin{abstract} 
\textcolor{black}{
This paper presents Parallel ARC (P-ARC), \textcolor{black}{a parallel formulation of the Adaptive Robot Coordination (ARC) approach to multi-robot motion planning (MRMP) which exploits subproblem independence.} 
ARC’s adaptive (de)composition of the multi-robot planning space exposes parallelism: single-robot paths are solved independently and iterative conflict detection and resolution create locally coupled subproblems. 
While distributing single-robot queries is trivial, not all conflicts are independent, so P-ARC proposes robot-disjoint conflict batches which enable efficient distributed detection and concurrent repair.
Additionally, OR-multi-start strategies are employed at the global and subproblem resolution levels, creating a hybrid parallel strategy OR-P-ARC.
We evaluate the methods against
sequential ARC, multi-start OR-ARC, and coupled and prioritized parallel baselines on controlled 2D mobile robot and planar-manipulator problems with
up to 256 robots and 3D Panda manipulator problems with up to 16 robots. 
On 16-robot Panda tasks, with 16 workers,
P-ARC and OR-P-ARC achieve 3.48× and 6.67× speedups, respectively. 
}
\end{abstract}


\section{Introduction}
\label{sec:introduction}

\textcolor{black}{
Planning collision-free motions for large robot teams remains a central challenge for deploying robotic solutions.
These motions must be computed quickly for multi-robot systems to operate responsively in real-world environments.
However, planning becomes increasingly difficult as more robots must be coordinated.
}

\textcolor{black}{
Coupled methods preserve coordination by directly searching the composite configuration space, but the exponential growth of that space with the team size limits scalability.
Decoupled methods, typically prioritized approaches, reduce the expansion of the search space by planning for robots individually but often sacrifice coordination or completeness guarantees.
Hybrid methods attempt to balance these tradeoffs by dynamically coupling where coordination is needed.
}

\textcolor{black}{
As modern processors provide increasing numbers of CPU cores, parallelism becomes an important avenue for accelerating planning.
Existing approaches expose parallelism at several levels.
Hardware-oriented approaches~\cite{bialkowski2011massively,thomason2024motions,motes2026multi} accelerate collision checking and motion validation,
shared-search methods distribute work within a single search process~\cite{carpin2002parallel,guo2025targeted}, 
OR-parallel methods exploit runtime variability through multiple independent searches~\cite{carpin2002parallel,ribeiro2025parallel},
and decomposition methods solve independent subproblems concurrently~\cite{lmma-phccsfomp-21}.
Each use of parallelism \textcolor{black}{identifies operations it can run independently in parallel.
}}

\begin{figure}[t]
    \centering
    \hspace{-1em}
    \begin{subfigure}[t]{0.25\textwidth}
        \centering
        \includegraphics[width=\linewidth]{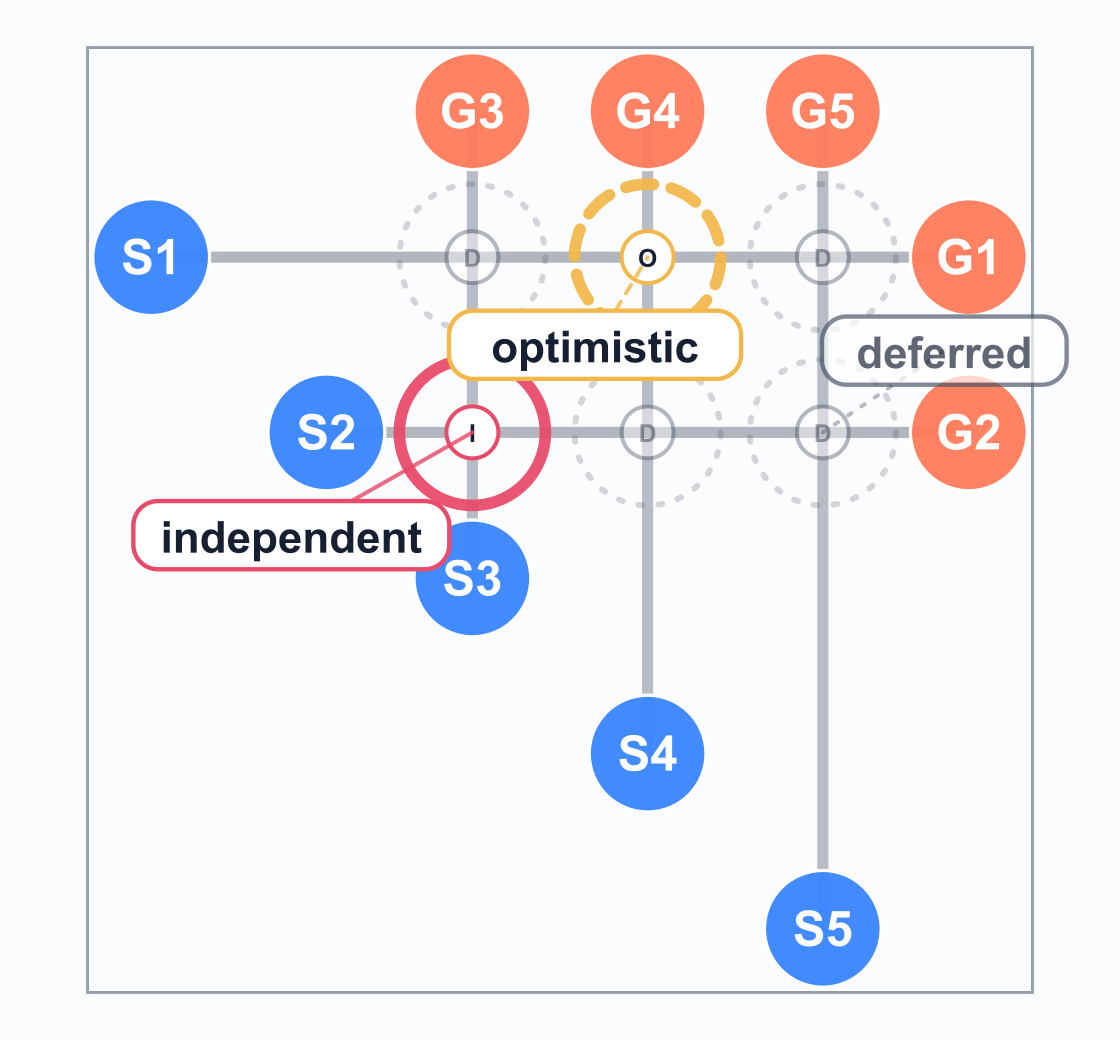}
        \caption{Conflict Dependence}
        \label{fig:conflict-dependence}
    \end{subfigure}
    \hspace{-1em}
    \begin{subfigure}[t]{0.25\textwidth}
        \centering
        \includegraphics[width=1.0\linewidth]{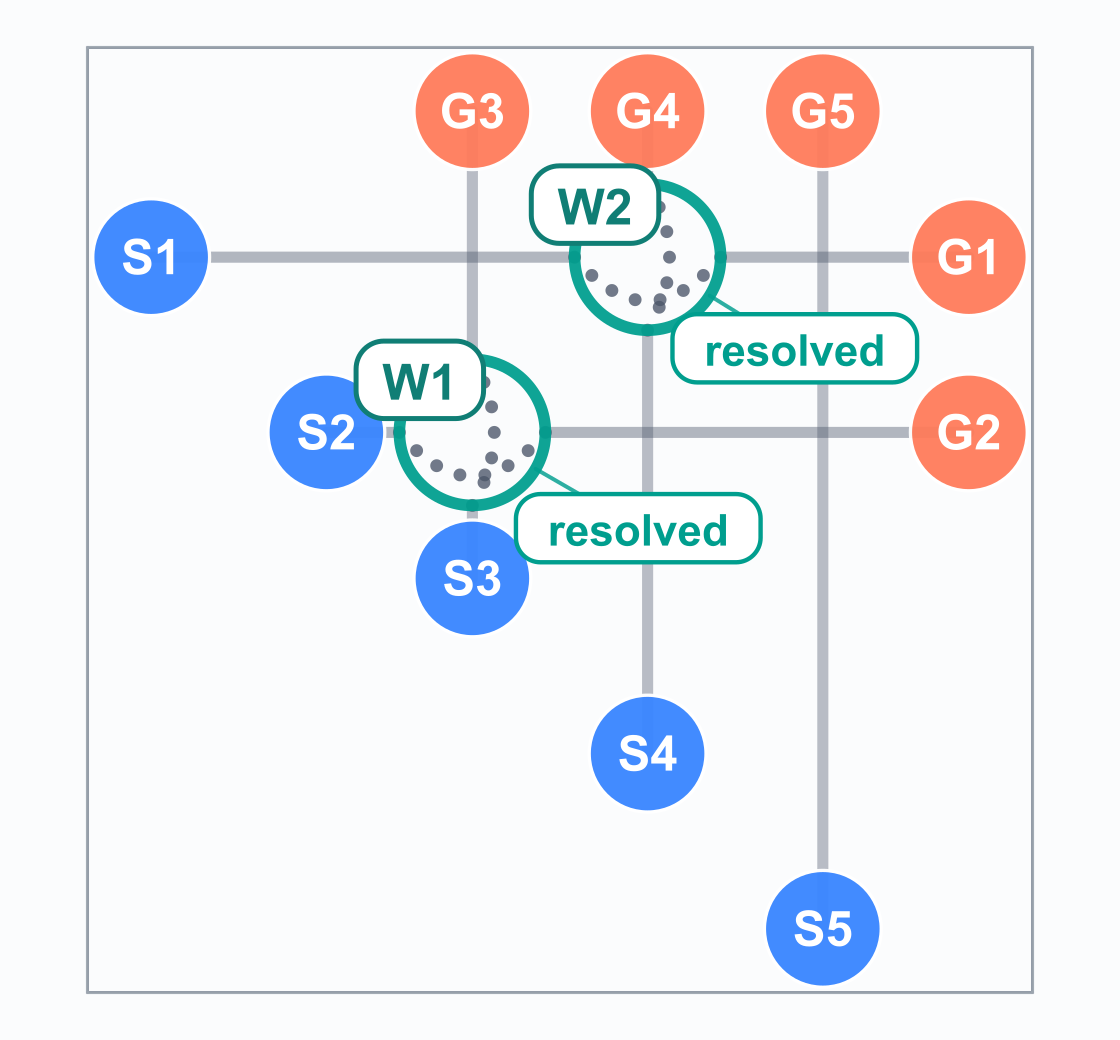}
        \caption{Conflict Resolution}
        \label{fig:conflict-resolution}
    \end{subfigure}
    \hspace{-1em}
    \caption{
    \small
    \textcolor{black}{
    Blue and orange circles denote robot starts and goals; 
    gray lines are individual paths. 
    (a) {\parc} selects a batch of robot-disjoint conflicts to resolve concurrently (red and yellow rings).
    The red solid ring depicts an independent conflict: it is the earliest conflict in either robot's path.
    The yellow dashed ring depicts an optimistically independent conflict.
    It is optimistic because the repair of the red conflict is expected to eliminate the earlier gray conflicts that currently prevent the yellow conflict from being earliest for its robots.
    The remaining conflicts (gray rings) are deferred as the timing of the path of at least one robot in each conflict is likely to be modified by resolutions of earlier-occurring conflicts, potentially removing or changing these conflicts.
    (b) The two concurrent repairs, assigned to workers $W_1$ and $W_2$ (green regions and dotted path segments), change path timing and eliminate the deferred conflicts.
     }
    }
    \label{fig:conflict-independence}
    \vspace{-2em}
\end{figure}

\textcolor{black}{
Adaptive Robot Coordination ({\arc})~\cite{solis2024adaptive} is a hybrid {\mrmp} approach which \textcolor{black}{dynamically creates (potentially) independent subproblems}: it first plans for each robot independently before iteratively searching for conflicts and resolving them by solving locally coupled subproblems around conflicts.
This dynamic local coupling avoids the complexity of the composite space while enabling necessary coordination; however, it makes parallelization nontrivial as conflicts are often dependent on the resolution of earlier conflicts (Fig.~\ref{fig:conflict-independence}). 
Efforts to parallelize {\arc} must therefore efficiently identify which conflicts can be processed concurrently.
}

\textcolor{black}{
We present Parallel {\arc} ({\parc}), an independence-aware parallel formulation of {\arc} \textcolor{black}{which identifies and exploits the dynamically exposed independent subproblems with parallelism.}
{\parc} distributes initial single-robot planning, parallelizes detection of robot-disjoint conflict batches, and repairs the conflicts in each batch concurrently. 
When few repairs are available, idle workers perform local OR-multi-start attempts on repair problems.
We also employ a global OR-multi-start strategy to create {\orparc}, dividing a fixed worker budget among independent instances of {\parc}, thereby combining subproblem parallelism with runtime diversity.
\textcolor{black}{
We evaluate {\parc}, {\orarc}, and a multi-start {\orarc} implementation against coupled and prioritized parallel baselines.
We study scaling with worker count and robot-team size on controlled 2D mobile robot and planar manipulator scenarios with up to 256 robots, along with realistic 3D Panda multi-manipulator planning problems with up to 16 robots, demonstrating over a 6$\times$ speedup over {\arc} on this hardest problem.
}}

\begin{table*}[t]
\centering
\caption{Landscape of Parallel Methods for Multi-Robot Motion Planning}
\small
\setlength{\tabcolsep}{0pt}
\renewcommand{\arraystretch}{1.35}
\definecolor{parcRule}{HTML}{4A5568}
\definecolor{parcPaletteBlue}{HTML}{3A86FF}
\definecolor{parcPaletteTeal}{HTML}{009E8E}
\definecolor{parcPaletteOrange}{HTML}{FF7A59}
\definecolor{parcPalettePurple}{HTML}{7C3AC5}
\definecolor{parcPaletteGold}{HTML}{F2B84B}
\colorlet{parcValidationFill}{parcPaletteBlue!22!white}
\colorlet{parcSearchFill}{parcPaletteTeal!18!white}
\colorlet{parcContributionsFill}{parcPaletteOrange!24!white}
\colorlet{parcProblemFill}{parcPaletteGold!28!white}
\colorlet{parcParallelFill}{parcPalettePurple!18!white}
\definecolor{parcNeutralFill}{HTML}{F5F5F5}
\tikzset{%
  parcColumnGroup/.style={rounded corners=1.5pt, draw=parcRule, line width=0.55pt, fill=#1},%
  parcRowGroup/.style={rounded corners=1.5pt, draw=parcRule, line width=0.55pt, fill=#1}%
}
\begin{NiceTabular}{@{}>{\centering\arraybackslash}m{0.70cm}>{\raggedright\arraybackslash}m{2.00cm}*{11}{>{\centering\arraybackslash}m{1.20cm}}@{}}
\CodeBefore
  \tikz \draw [parcColumnGroup=parcValidationFill] (1-|3) rectangle (3-|6);
  \tikz \draw [parcColumnGroup=parcSearchFill] (1-|6) rectangle (3-|11);
  \tikz \draw [parcColumnGroup=parcContributionsFill] (1-|11) rectangle (3-|last);
  \tikz \draw [parcRowGroup=parcProblemFill] (3-|1) rectangle (5-|3);
  \tikz \draw [parcRowGroup=parcParallelFill] (5-|1) rectangle (9-|3);
\Body
 &  & \Block{1-3}{\bfseries Validation} &  &  & \Block{1-5}{\bfseries Search} &  &  &  &  & \Block{1-3}{\bfseries Contributions (Ours)} &  &  \\
 &  & {\scriptsize\makecell{GPU-CC-\\RRTs~\cite{bialkowski2011massively}}} & {\scriptsize\makecell{VAMP~\cite{thomason2024motions}}} & {\scriptsize\makecell{VA-\\MRMP~\cite{motes2026multi}}} & {\scriptsize\makecell{PHC-\\CBS~\cite{lmma-phccsfomp-21}}} & {\scriptsize\makecell{DP/PB-\\ECBS~\cite{guo2025targeted}}} & {\scriptsize\makecell{EP-\\RRT~\cite{carpin2002parallel}}} & {\scriptsize\makecell{OR-EP-\\RRT~\cite{carpin2002parallel}}} & {\scriptsize\makecell{Multi-\\Threaded\\TEA*\cite{ribeiro2025parallel}}} & {\scriptsize P-ARC} & {\scriptsize OR-ARC} & {\scriptsize OR-P-ARC} \\
\Block{2-1}{\rotatebox[origin=c]{90}{\scriptsize\bfseries Problem}} & Multi-robot &  &  & \ensuremath{\checkmark} & \ensuremath{\checkmark} & \ensuremath{\checkmark} & \ensuremath{\checkmark} & \ensuremath{\checkmark} & \ensuremath{\checkmark} & \ensuremath{\checkmark} & \ensuremath{\checkmark} & \ensuremath{\checkmark} \\
 & State Space & {\cspace} & {\cspace} & {\cspace} & Grid & Grid & {\cspace} & {\cspace} & Grid & {\cspace} & {\cspace} & {\cspace} \\
\Block{4-1}{\rotatebox[origin=c]{90}{\scriptsize\bfseries \makecell{Use of\\Parallelism}}} & Validation & \ensuremath{\checkmark} & \ensuremath{\checkmark} & \ensuremath{\checkmark} &  &  &  &  &  & \ensuremath{\checkmark} &  & \ensuremath{\checkmark} \\
 & Subproblem &  &  &  & \ensuremath{\checkmark} &  &  &  &  & \ensuremath{\checkmark} &  & \ensuremath{\checkmark} \\
 & Shared Search &  &  &  &  & \ensuremath{\checkmark} & \ensuremath{\checkmark} & \ensuremath{\checkmark} &  &  &  &  \\
 & OR-Multi-Start &  &  &  &  &  & & \ensuremath{\checkmark} & \ensuremath{\checkmark} & (Local) & \ensuremath{\checkmark} & \ensuremath{\checkmark} \\
\end{NiceTabular}
\label{tab:parc-related-work}
\end{table*}

\textcolor{black}{
In summary, our contributions are:
\begin{itemize}
    \item {\parc}, an independence-aware parallel formulation of ARC for evolving robot couplings
    \item Efficient distributed detection of robot-disjoint conflict batches
    \item \textcolor{black}{{\orparc}}, a hybrid subproblem and OR-multi-start parallelism \textcolor{black}{which demonstrates better success rates, faster planning times, and higher quality solutions on difficult planning problems}
    \item \textcolor{black}{Evaluation of both methods across domains where they are well and poorly suited to fully characterize behavior}
\end{itemize}
}

\section{Background and Related Work}
\label{sec:related-work}

In this section, we define the motion planning problem and discuss sequential and parallel approaches to it.

\subsection{Multi-Robot Motion Planning}

Multi-robot motion planning ({\mrmp}) seeks a continuous path $p$ for each robot $r\in R$ from its start position to its goal position.
Each path is time parameterized such that a timestep $t$ defines a configuration of $r$ along $p$.
Two paths contain a conflict $c=\{r_i,r_j,t\}$ if the configurations of $r_i$ and $r_j$ at $t$ along $p_i$ and $p_j$ are in collision.
A set of paths forms a valid solution if (i) no individual path has a collision with any obstacles and (ii) there are no conflicts between any paths.

{\mrmp} formally searches the composite configuration space {\compcspace} or the Cartesian product of the individual {\cspaces} of the set of robots.
However, the size of this space increases exponentially with the number of robots, leading to two approaches to the {\mrmp} problem.
Decoupled methods plan for robots independently, typically in a prioritized manner treating higher priority robots as dynamic obstacles~\cite{van2005prioritized,kerimov2025si}.
This reduces the complexity of the search space, often improving planning time, but sacrifices completeness guarantees and coordination potential.
Coupled methods directly search {\compcspace}, accepting the size of the space in order to achieve higher levels of coordination~\cite{sl-uppccdpmrs-2002,ssh-faniaehdrfeoirimm-16,ssdhb-dsaiammp-20}.
Hybrid methods attempt to balance the tradeoffs of coupled and decoupled approaches~\cite{solis2024adaptive,wc-sefmpp-15}.
These typically start by decoupling the search and then dynamically choose when to couple robots.

\subsection{Parallel Approaches}

\textcolor{black}{
Table I provides an overview of the landscape of parallelism in multi-robot planning.
The two main categories are parallel validation and parallel search.
GPU~\cite{bialkowski2011massively} and CPU SIMD~\cite{thomason2024motions} parallel validation approaches have been developed for single robot planning. 
Recent work~\cite{motes2026multi} has extended SIMD validation to multi-robot-specific planning.
By using hardware-based parallelism, these are often compatible with search parallelism as is demonstrated in our experiments.
}

\textcolor{black}{
Parallel search can be further categorized into subproblem, shared-search, or OR-multi-start approaches.
Subproblem approaches decompose the {\mrmp} problem into (assumed) independent subproblems, solve them in parallel, and then reconcile the solutions against each other. 
PHC-CBS~\cite{lmma-phccsfomp-21} does this for grid-world MAPF problems by hierarchically planning for and then merging subsets of robots until recovering a global solution.
}

\textcolor{black}{
Several non-parallel {\mrmp}  methods decompose problems into subproblems~\cite{van2005prioritized,zanardi2023factorization,solis2024adaptive}. 
The relative independence of these subproblems determines their suitability for parallelism. 
For example, prioritized planning plans for each robot individually, but it treats higher priority robots as dynamic obstacles, which forces sequential planning of each robot in the priority list instead of solving them in parallel. 
The temporal and spatial decomposition in~\cite{guo2021spatial} chooses intermediate subgoals for each robot, creating independent smaller problems which can be solved in parallel.
However, it still considers all robots within each segment and forces synchronization of the whole team from segment to segment.
ARC~\cite{solis2024adaptive} assumes independence and plans for robots or subsets of robots completely ignoring the rest of the team (details in Section~\ref{sec:arc}).
The factorization approach in \cite{zanardi2023factorization} starts coupled and searches for states where it can branch the search into independent threads, using a hypergraph to model this in one data structure.
This single data structure approach and multiple possible instances of decompositions of the same robot subsets present a challenge for how and when to spin off and reconcile subproblems, though it is well suited to the shared-search parallelism used by methods like parallel RRT~\cite{carpin2002parallel}. 
}

\textcolor{black}{
Shared-search methods~\cite{carpin2002parallel,guo2025targeted} find ways to parallelize work within a single search process.
DP/PB-ECBS~\cite{guo2025targeted}, another grid-world MAPF method, uses a decentralized approach to search separate branches of its conflict tree, and EP-RRT~\cite{carpin2002parallel} grows a single search tree with multiple workers. 
}

\textcolor{black}{
Finally, OR-multi-start parallelism exploits the observation that many of these methods are randomized which can lead to significant variance in their runtimes. 
EP-RRT~\cite{carpin2002parallel} exploits this by starting many runs in parallel with different random seeds for its RRT search. 
The prioritized planning approach~\cite{ribeiro2025parallel} recognizes the dependence of planning success and speed on priority order and runs different orders in parallel.
We adapt this to sampling-based planning with ST-RRT~\cite{grothe2022st} as the underlying planner to create OR-PP- ST-RRT to use as a parallel baseline along with an adaptation of \cite{carpin2002parallel} to RRT-C~\cite{kuffner2000rrt}. 
}

\subsection{Adaptive Robot Coordination ({\arc})}
\label{sec:arc}

ARC~\cite{solis2024adaptive} consists of three stages: (i) initial solutions, (ii) conflict detection, and (iii) conflict resolution.
In stage (i), the method computes an independent path for each robot in its own {\cspace}, ignoring robot-robot collisions.
Stage (ii) then performs a linear scan along these independent paths, checking each timestep for any robot-robot collision, returning the first occurring conflict defined by the robot pair and timestep $c=\{r_i,r_j,t\}$ along paths $p_i,p_j\in P$.

Stage (iii) takes this conflict and creates a subproblem.
Local start and goal configurations are taken from the independent paths a given number of timesteps before and after the discovered conflict and a local {\cspace} is defined for each robot containing the respective path segment (with some buffering).
It then employs a configurable hierarchy of {\mrmp} strategies to solve the subproblem.
If all {\mrmp} methods fail, then the time horizon is expanded, and a new subproblem is created and attempted.
In the worst case, the subproblem expands to the robots' global starts and goals and the entire {\cspace} is considered for the involved robots.

In addition to time horizon expansion, the subproblems can expand to include more robots.
Prior subproblem solution time windows are cached.
If a conflict $c=\{r_i,r_j,t\}$ is discovered and there is a prior resolution between $r_i$ and $r_k$ that contains $t$, then the initial conflict is expanded to include $r_k$ before generating the subproblem.
This aims to keep synchronization between robots during portions of their paths that have been found to require coordination.

A successful subproblem solution is used to replace the conflicting segments of the involved robots.
The segments are not required to be the same length as the original, so the time synchronization of the path after the patch may shift in either direction.
As such, {\arc} resolves conflicts in order of their occurrence in the paths.
This avoids wasting effort on a conflict late in a path that might be naturally shifted when resolving a conflict earlier in the path (Fig.~\ref{fig:conflict-independence}).
Stages (ii) and (iii) are repeated until no conflicts remain.

\section{Parallel ARC ({\parc})}
\label{sec:method}

In this section, we propose parallel approaches to each stage of ARC to create {\parc} and discuss the theoretical properties of {\parc}.
In addition to the stage-level parallelism, we consider a global OR-multi-start strategy which runs multiple instances of {\parc} in parallel and accepts the first solution discovered. 
Given a budget of $W$ available workers, we distribute them among $K$ instances of {\parc} such that each instance has $W_k$ workers available and $\sum_{k=1}^{K} W_k=W$.

The following subsections detail how {\parc} employs $W_k$ worker processes within each stage.
We use a simple fork-join execution model at each stage (though this is not a required implementation choice).
A central coordinator process maintains the set of robot paths.
At the beginning of each stage, the coordinator forks $W_k$ worker processes.
Workers inherit the planning problem and current solution state, execute the assigned tasks, and return the results to the coordinator.
The coordinator synchronizes the workers at the end of each stage, reducing the set of conflicts or updating the set of global paths, and then proceeds to the next stage.

\subsection{Initial Solutions}
\label{sec:method:initial}

{\arc} begins by planning an individual path for each robot, independent of all other robots.
This is the most straightforward stage to parallelize.
Given $N$ robots and $W_k$ workers, initial motion planning tasks are distributed among the $W_k$ workers.
When a worker completes its current task, it returns the solution to the main process and another motion planning task is \textcolor{black}{assigned} to it.
As the set of individual robot motion tasks may vary in difficulty and planning time, this helps distribute the actual computational load rather than the robot load.
If there are more workers than remaining individual motion tasks, the OR-multi-start strategy can be applied locally to long-running tasks.
\textcolor{black}{Alternatively, extra workers could be used to parallelize the work within a single solve attempt if the solver is a parallel search method like EP-RRT~\cite{carpin2002parallel}, but that is outside the scope of this work.}

\subsection{Conflict Detection}
\label{sec:method:conflict-dection}

\textcolor{black}{
Given the current set of paths $P$, the conflict detection stage returns a batch of conflicts $B$ to be resolved in parallel in the conflict resolution stage.
The sequential search for conflicts initializes a conflict batch $B=\emptyset$ and scans for conflicts linearly from the first to the last timestep in $P$.
A conflict $c$ is added if its robot set is disjoint from the robot sets already in $B$.
In {\arc}~\cite{solis2024adaptive}, the subproblem for a conflict $c=\{r_i,r_j,t\}$ is expanded from $r_i,r_j$ to the set of robots $R(c)$ which have previous resolutions involving $r_i$ or $r_j$ containing the timestep $t$. 
To ensure the conflicts included in $B$ are robot-disjoint, this robot expansion mechanism is now performed upon discovery of a conflict and is used to determine whether a new conflict $c$ is disjoint from the robot sets already in $B$.
The search stops when $|B|=W_k$ or the scan reaches the last timestep.
}

\subsubsection{Optimistically Independent Conflicts}

\textcolor{black}{
The requirement of disjoint robot sets for conflicts in $B$ results in the inclusion of two types of conflicts: independent and \textit{optimistically independent}.
Given all conflicts $C$ in $P$, a conflict $c=\{r_i,r_j,t\}$ is independent if there exists no $c'=\{r_i',r_j',t'\}\in C, c'\neq c, R(c)\cap R(c')\neq\emptyset, t'<t$.
That is to say, $c$ is the earliest occurring conflict in $C$ involving any robots in $R(c)$.
Limiting parallel resolution to only independent conflicts ensures that no two parallel resolutions modify the same individual robot paths. Concurrent conflict resolutions involving the same robot can introduce discontinuities if the repair windows overlap.
Even non-overlapping repairs may desynchronize repairs for later-occurring conflicts potentially reintroducing similar conflicts and wasting work.
This desynchronization, a consequence of repair path segments varying in duration from the original conflict segment, can also implicitly resolve later-occurring conflicts as robots may arrive earlier or later to the location of an original conflict (Fig.~\ref{fig:conflict-independence}).
}

\textcolor{black}{
As this desynchronization almost always happens, we introduce \textit{optimistically} independent conflicts.
Optimistically independent conflicts are those which would become independent if the resolutions of earlier independent conflicts resolve (via desynchronization) later-occurring conflicts for involved robots (e.g., the yellow circle in Fig.~\ref{fig:conflict-dependence}).
This definition is recursive, where earlier optimistically independent conflicts are treated as independent when determining the optimistic independence of later conflicts.
More formally, a conflict $c_\text{opt}\in C$ is optimistically independent if every earlier conflict $c'\in C$ with nonempty shared set $R_\text{shared}=R(c_\text{opt})\cap R(c')\neq\emptyset$, and, for each $r_k\in R(c')\setminus R_\text{shared}$, there exists an independent or optimistically independent conflict $c''_k\in C$ involving $r_k$ that occurs before $c'$.
}

\textcolor{black}{
While optimistically independent conflicts may lead to redundant or wasted effort, this rarely happens in practice; instead it increases the number of conflicts available for concurrent repairs.
Additionally, strict conflict independence requires collision checking each pair unless both have been involved in an earlier conflict to ensure independence.
For example, in Fig.~\ref{fig:conflict-independence}, enforcing strict independence requires discovering the $r_2,r_4$ and/or the $r_1,r_3$ conflicts to know that the yellow $r_1,r_4$ conflict is not strictly independent.
Accepting optimistically independent conflicts allows the search to stop collision checking a robot pair once either robot is involved in a conflict added to the batch.
The impact of accepting optimistically independent conflicts is studied in Section~\ref{sec:ablations-optimistic-independence}.
}

\subsubsection{Parallel Conflict Search}

\textcolor{black}{
Robot pairs are distributed uniformly among the workers in a round-robin manner.
Each worker scans its assigned pairs for conflicts.
Distributed conflict checking reduces the wall-clock time to check the entire timestep range for conflicts. 
However, it also eliminates an early-termination property of sequential conflict detection, since the first $W_k$ (optimistically) independent conflicts may occur on different workers.
}

\textcolor{black}{
To recover this, we introduce a synchronization horizon $H$.
Each worker searches its assigned robot pairs for the next $H$ timesteps before reporting candidate conflicts to the coordinator.
The coordinator merges candidate conflicts in temporal order, selects the earliest robot-disjoint (optimistically independent) conflicts to add to $B$, and either terminates once $|B|=W_k$ or advances all workers to the next synchronization horizon.
}

\textcolor{black}{
To further reduce redundant computation, the last validated timestep for every robot pair is cached across conflict detection iterations.
Conflict checking resumes from this cached timestep unless either robot participates in a repair occurring earlier along its path, in which case the cached timestep is reset to the earliest modified timestep.
Similarly, once a robot has been included in a conflict in the current batch, subsequent conflict checks involving that robot are skipped during the remainder of that conflict detection iteration.
This skip information is propagated between workers at each synchronization horizon.
}

\subsection{Conflict Resolution}
\label{sec:method:conflict-resolution}

Given a batch of \textcolor{black}{(optimistically)} independent conflicts $B$, each $c\in B$ can be given to a worker and solved independently.
Each subproblem resolution returns the patch to the paths for the involved robots.
{\parc} waits on all subproblems to finish before checking for conflicts again.

Like when planning initial solutions, if there are more workers than remaining subproblems, a local OR-multi-start strategy is employed.
\textcolor{black}{
Workers are continuously assigned to unsolved subproblems in a round-robin manner as the workers become available.
When a subproblem is solved, all remaining instances of it are canceled and the workers reassigned.
}
This keeps our CPU utilization high, even when there are few \textcolor{black}{(optimistically)} independent conflicts or \textcolor{black}{extreme variance in subproblem difficulty.}

\subsection{Theoretical Properties}
\label{sec:method:theoretical}

{\arc}~\cite{solis2024adaptive} maintains the completeness guarantees of the strongest method in its {\mrmp} solver hierarchy.
The original {\arc} paper also discusses including a probabilistically complete method as the final attempt once a subproblem has reached the global starts and goals for each involved robot to ensure probabilistic completeness for the method~\cite{solis2024adaptive}.
We maintain these properties by (i) finding all conflicts and (ii) expanding resolutions to the global problem if necessary.

The distribution of conflict detection work described in Section ~\ref{sec:method:conflict-dection} ensures that any returned solution has been checked for (and not found any) conflicts between all pairs of robots at all timesteps.
If at least one conflict exists, the conflict detection routine will find it.
The subproblem resolution mechanism for individual conflicts is the same in {\parc} as it is in {\arc}.
\textcolor{black}{
Each unresolved subproblem is repeatedly expanded, eventually reaching the global scope if necessary.
Under the same assumptions as {\arc} and with unbounded sampling time, the probability of finding a solution approaches one.
}
The OR-multi-start strategies at both the global and local levels make no modification to the methods run, and thus do not impact the completeness properties of the individual attempts or the global algorithm.

{\arc} makes no claim about solution quality as the local nature of the subproblem-based conflict resolution ignores potentially better alternatives outside the local region.
We maintain this and, empirically, do not see any degradation in solution quality from {\arc} to {\parc}, though we do observe that OR-multi-start sometimes improves solution quality.

\section{Evaluation}
\label{sec:evaluation}
We conduct a set of experiments to evaluate the impact of parallelization on {\arc}.
Section~\ref{sec:experimental-design} details our experimental design.
Section~\ref{sec:results} presents and discusses our results.

\subsection{Experimental Design}
\label{sec:experimental-design}

The experiments are designed to evaluate (i) whether {\parc} improves runtimes with increasing numbers of workers, (ii) the behavior of the three stages of {\arc} as more workers are available, (iii) the impact of the OR-parallel multi-start strategy, and (iv) the impact of {\parc} in real-world-inspired settings.

For (i) and (ii), we design three 2D scenarios where we can control for conflicts and the associated work in conflict detection and conflict resolution (Fig.~\ref{fig:scenarios}).
In the scenarios with more difficult conflict resolutions, we include OR-multi-start parallelism to evaluate (iii), including different distributions of workers between OR-level and task-level parallelism.
These we denote {\orparc}-$K\times W_k$ as $K$ independent OR-parallel {\parc} instances, each using $W_k$ workers, for a total budget of $K\times W_k=W$ workers.

For (iv), we evaluate the different algorithm configurations on randomly generated tasks in the Panda Cage environment (Fig.~\ref{fig:scenarios-panda}).
This is intended to resemble large numbers of manipulators operating in a shared space in close proximity.
\textcolor{black}{
Finally, a pair of ablation studies is included to study the impact of including optimistically independent conflicts and different synchronization horizon lengths for distributed conflict detection. 
}

\subsubsection{\textcolor{black}{Scenarios}}
\label{sec:scenarios}

\begin{figure*}[t]
    \centering
    \begin{subfigure}[t]{0.14\textwidth}
        \centering
        \includegraphics[width=\linewidth]{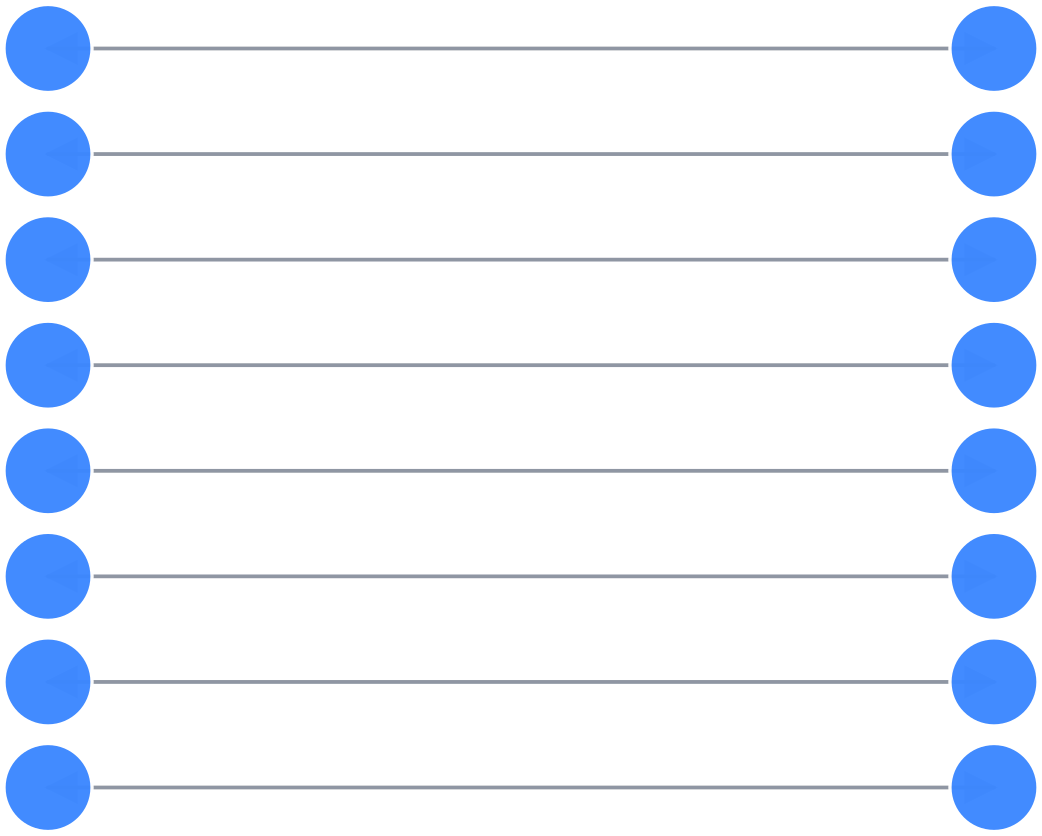}
        \caption{Mobile Cross}
        \label{fig:scenarios-2d-cross}
    \end{subfigure}
    \hfill
    \begin{subfigure}[t]{0.13\textwidth}
        \centering
        \includegraphics[width=1.0\linewidth]{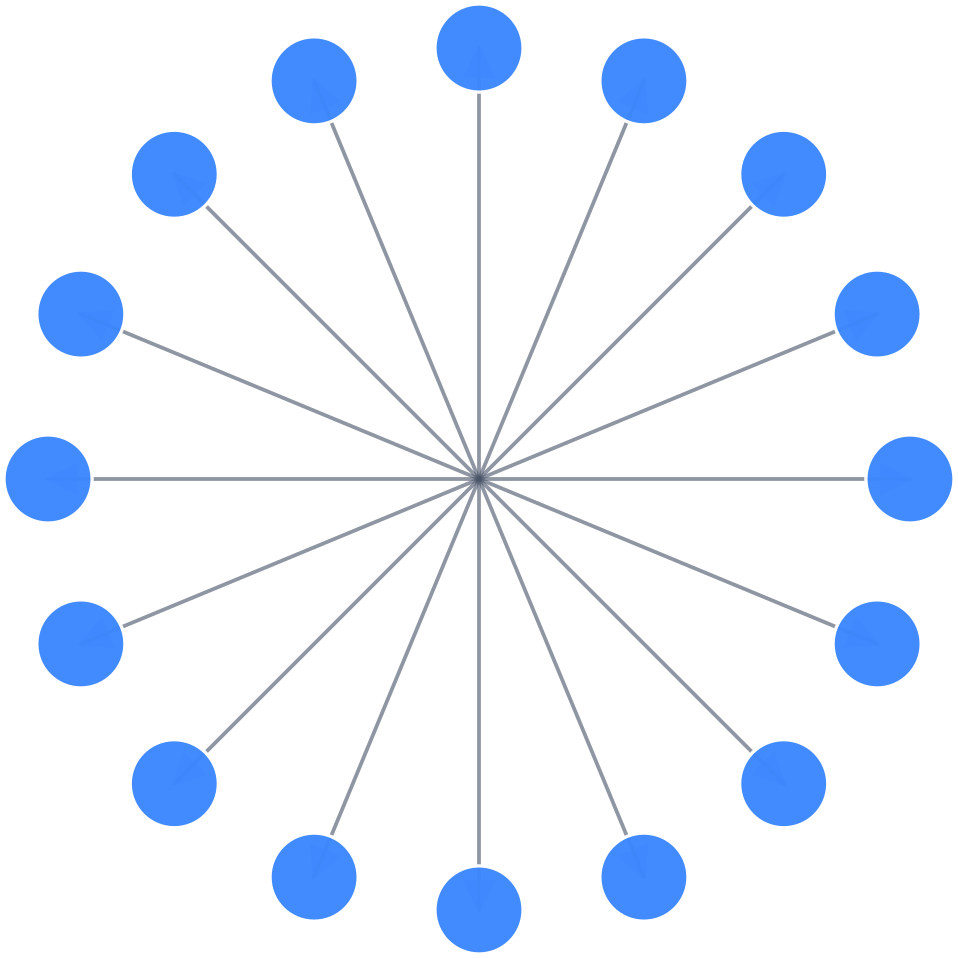}
        \caption{Mobile Circle}
        \label{fig:scenarios-2d-circle}
    \end{subfigure}
    \hfill
    \begin{subfigure}[t]{0.4\textwidth}
        \centering
        \includegraphics[width=1.0\linewidth]{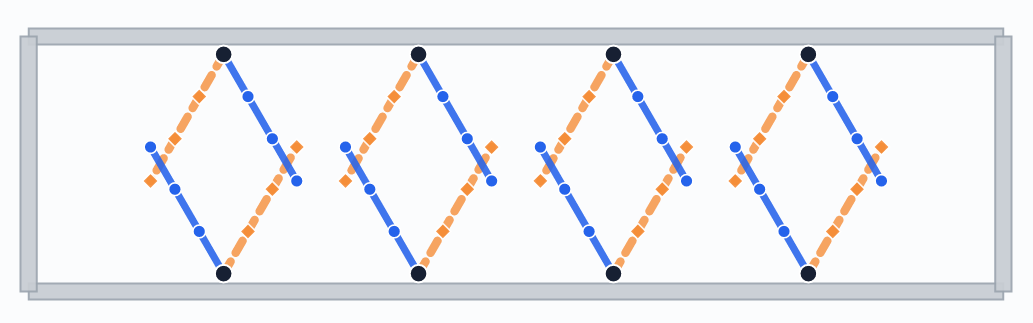}
        \caption{Planar Cross}
        \label{fig:scenarios-planar-cross}
    \end{subfigure}
    \hfill
    \begin{subfigure}[t]{0.2\textwidth}
        \centering
        \includegraphics[width=1.0\linewidth]{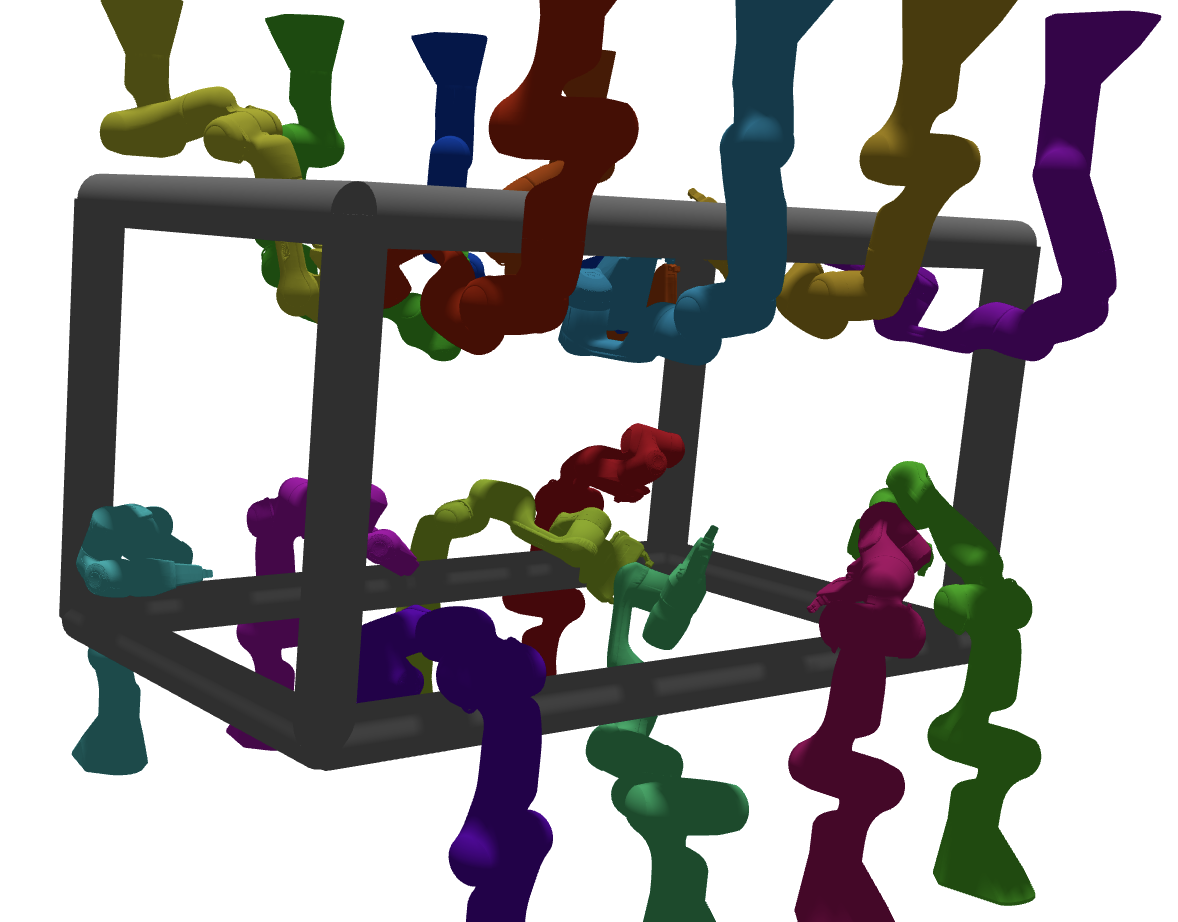}
        \caption{Panda Cage}
        \label{fig:scenarios-panda}
    \end{subfigure}
    \caption{
    \small
    (a) The Mobile Cross scenario has two parallel lines of robots swapping places, inducing conflicts between pairs of robots.
    (b) The Mobile Circle scenario has robots arranged in a circle that swap places with their diametrically opposed partners which causing pairwise conflicts to cascade into one large coupled subproblem involving all robots.
    (c) The Planar Cross scenario has each robot start in the blue configuration and move to the orange configuration inside the gray box, inducing pairwise conflicts.
    \textcolor{black}{(d) The Panda Cage scenario has manipulators mounted to the floor and ceiling and operating in a shared workspace resembling real-world scenarios like automobile manufacturing.}
    }
    \label{fig:scenarios}
\end{figure*}

\textcolor{black}{
We evaluate four scenarios.
In the three 2D scenarios, each planner is run for 30 trials with a 30 second timeout.
In the 3D Panda Cage scenario, each planner is run for 10 trials on each of the five tasks with a 100 second timeout.
}

\textbf{Mobile Cross:}
\textcolor{black}{2D mobile robots are arranged in} two parallel lines (Fig.~\ref{fig:scenarios-2d-cross}).
Each robot swaps places with the robot opposite it, creating $\frac{N}{2}$ conflicts for $N$ robots.
\textcolor{black}{Conflicts are almost exclusively independent. 
Resolutions are simple.
}

\textbf{Mobile Circle:}
This scenario (Fig.~\ref{fig:scenarios-2d-circle}) is designed to show when {\parc} performs poorly.
Robots swap places with the opposite partner in a circle.
This is intended to draw all robots into a single subproblem, reducing the independence of the individual robot paths and thus the effectiveness of parallel conflict resolution.

\textbf{Planar Cross:}
\textcolor{black}{
Two lines of planar manipulators must move from the blue to the orange configuration in Fig.~\ref{fig:scenarios-planar-cross}.
Increased robot complexity adds to the cost of conflict detection and subproblem resolution.
The constrained environment also often requires subproblems expanded to include additional robots to resolve conflicts (though not to the extreme of Mobile Circle). 
}

\textbf{Panda Cage:}
This scenario is intended to demonstrate the utility of {\parc}, {\orarc}, and {\orparc} for large numbers of robots in real-world-inspired settings.
\textcolor{black}{For each team size, five random tasks are generated with the end effector of each robot inside the cage at the start and goal positions.}

\subsubsection{Implementation and Baselines}

All {\arc} variants use a subproblem solver hierarchy with a cheap prioritized planning adaptation~\cite{kerimov2025si} of ST-RRT~\cite{grothe2022st} (PP-ST-RRT) \textcolor{black}{without rewiring and a composite RRT-C.}
Conflict window sizes and expansion rates are empirically tuned for {\arc} across all scenarios and used across the parallel variants.

\textcolor{black}{We include coupled and decoupled parallel {\mrmp} baselines.}
\textcolor{black}{For the coupled baseline, we adapt the embarrassingly parallel RRT (EP-RRT) {\mrmp} variants from~\cite{carpin2002parallel} to RRT-C~\cite{kuffner2000rrt}}.
This includes an internal parallel tree growth (EP-RRT-C), an OR-multi-start strategy applied to RRT-C (OR-RRT-C), and a mixed parallel OR-EP-RRT-C.
The same $K\times W_k$ notation is used to indicate the distribution of $W_k$ workers to each of the $K$ instances of OR-EP-RRT-C.
For the decoupled baseline, we use the OR-multi-start prioritized planner strategy~\cite{ribeiro2025parallel} (which samples different priority orders) with PP-ST-RRT~\cite{kerimov2025si} (OR-PP-ST-RRT).
We run these with 16 workers on each of the scenarios and report the runtime performance relative to the {\arc} variants.

\textcolor{black}{All implementations build on the vector-accelerated multi-robot motion planning (VA-MRMP) implementations from~\cite{motes2026multi} and get the {\simd} parallel validation for free.}
All experimental trials were run on a Linux workstation (x86\_64, kernel 6.8) equipped with an Intel Core i7-14700F CPU (up to 5.4 GHz) and 32 GB of RAM.
\textcolor{black}{\textit{Our implementations will be made open source upon acceptance of this paper to maintain a double-blind review.}}

\begin{figure*}[t]
    \centering
    \includegraphics[width=\linewidth]{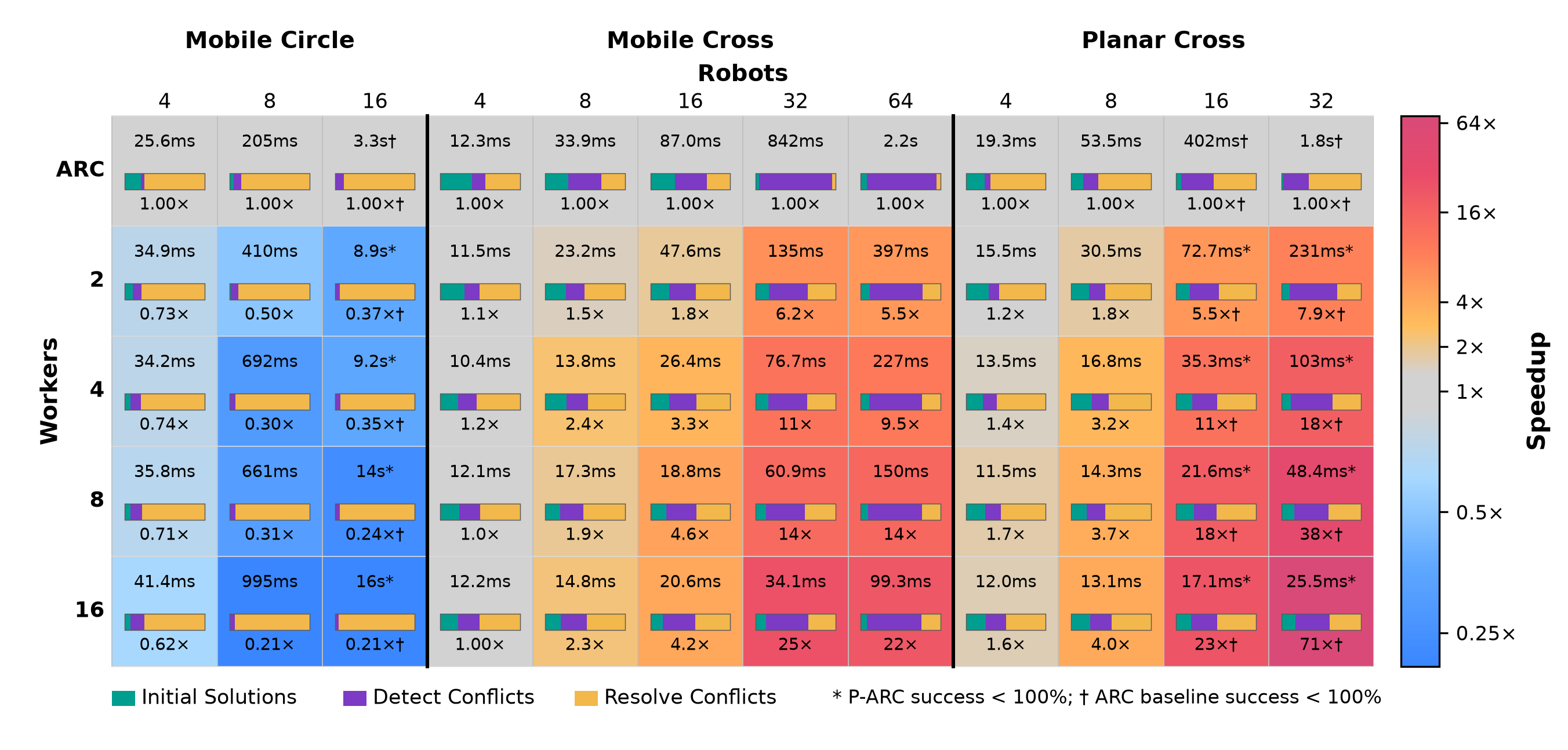}
    \caption{\textcolor{black}{
    Performance of {\parc} as the number of available workers and robots increase across all 2D scenarios.
    Data is reported for all scenarios in which {\arc} solved at least 50\% of the trials before the timeout.
    Values are computed from successful trials for both methods.
    The heatmap encodes median-runtime speedup compared to {\arc} on the same problem.
    Each cell contains (from top to bottom): median total runtime, time-per-stage breakdown, and speedup relative to {\arc}.
    The time-per-stage breakdown bar indicates what portion of the total runtime is allocated to each stage: initial solution, conflict detection, and conflict resolution.
    }}
    \label{fig:results:parallel-speedup}
\end{figure*}

\begin{figure*}[t]
    \centering
    \includegraphics[
        width=\textwidth
    ]{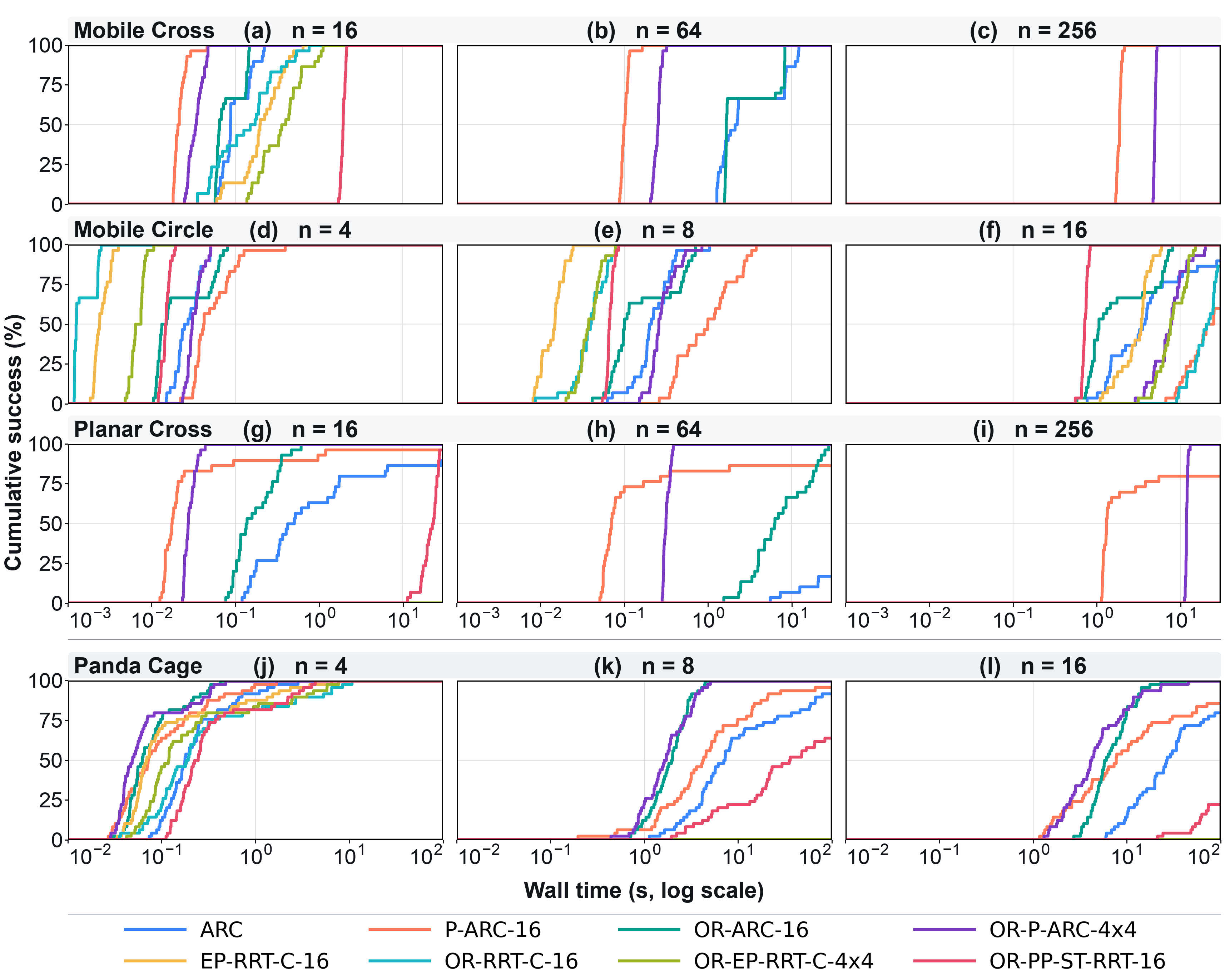}
    \caption{
    \textcolor{black}{
        Cumulative success as wall-clock time increases for all scenarios.
        Each panel header gives the number of robots \(n\).
        The 2D scenarios, Mobile Cross (a)--(c), Mobile Circle (d)--(f), and Planar Cross
        (g)--(i), report results over 30 random seeds and share a common x-axis scale.
        Panda Cage (j)--(l) report results for five random tasks run with 10 random seeds each.
        }
    }
    \label{fig:results}
\end{figure*}

\subsection{Results}
\label{sec:results}

\begin{table}[t]
\centering
\caption{Panda Cage Relative Speedups and Solution Costs}
\label{tab:panda-results}

\setlength{\tabcolsep}{3pt}
\renewcommand{\arraystretch}{1.1}

\resizebox{\columnwidth}{!}{%
\begin{tabular}{@{}lccc ccc@{}}
\toprule
\multirow{2}{*}{Method}
& \multicolumn{3}{c}{Runtime (s) / Speedup}
& \multicolumn{3}{c}{Makespan / Relative Cost} \\
\cmidrule(lr){2-4}
\cmidrule(l){5-7}
& 4 & 8 & 16 & 4 & 8 & 16 \\
\midrule
ARC & 0.18 & 7.18 & 26.83 & 3343 & 11170 & 11880 \\
\midrule
OR-ARC-16 & $2.68\times$ & $3.60\times$ & $4.50\times$ & $0.60\times$ & $0.57\times$ & $0.65\times$ \\
P-ARC-16 & $2.46\times$ & $1.69\times$ & $3.48\times$ & $0.94\times$ & $0.94\times$ & $0.79\times$ \\
OR-P-ARC-2$\times$8 & $3.19\times$ & $3.10\times$ & $6.32\times$ & $0.79\times$ & $0.78\times$ & $0.66\times$ \\
OR-P-ARC-4$\times$4 & {\boldmath\textbf{$3.64\times$}} & $4.30\times$ & $6.43\times$ & $0.64\times$ & $0.65\times$ & $0.70\times$ \\
OR-P-ARC-8$\times$2 & $3.59\times$ & {\boldmath\textbf{$6.02\times$}} & {\boldmath\textbf{$6.67\times$}} & {\boldmath\textbf{$0.58\times$}} & {\boldmath\textbf{$0.56\times$}} & {\boldmath\textbf{$0.61\times$}} \\
\bottomrule
\end{tabular}%
}

\vspace{2pt}
\begin{minipage}{\columnwidth}
\footnotesize
\raggedright
ARC reports median runtime (seconds) and median makespan (timesteps) baselines.
Other rows report speedup over ARC and relative solution cost.
Bold values are the best in each column.
\end{minipage}
\end{table}

\textcolor{black}{
{\parc} speedup details are provided in Fig.~\ref{fig:results:parallel-speedup} and Table~\ref{tab:panda-results}.
Runtime results for all methods with 16 workers are reported in Fig.~\ref{fig:results}.
}

\subsubsection{P-ARC Speedups}
\textcolor{black}{
We run {\parc} with increasing available workers and report the total runtime, time-per-stage, and speedup relative to {\arc} for each of the 2D scenarios with increasing robot team sizes.
The heatmap in Fig.~\ref{fig:results:parallel-speedup} shows the general trend of greater speedups as both the number of workers and the number of robots increase.
This supports our claim that, for problems with subproblem independence, {\parc} performance improves as more workers are available and that it scales to larger team sizes better than {\arc}.
Mobile Circle is the expected exception to this where there is little independence between conflicts, and the performance trends are reversed.
}

\textcolor{black}{
The time-per-stage breakdown in each cell of Fig.~\ref{fig:results:parallel-speedup} shows how the runtime allocation changes by problem type and number of robots, justifying the parallelization of each stage. 
Within each scenario, the runtime allocation is relatively unaffected by the available workers but changes with the team size.
Mobile Circle devolves into one expensive repair problem which leaves little conflict detection work.
The other 2D scenarios see conflict detection account for an increasing portion of the runtime as the team size increases. 
This is expected as conflict detection scales quadratically with the number of robots while the initial solutions and induced conflicts scale linearly in these scenarios.
}

\textcolor{black}{
We note that the greatest speedup, 71$\times$, comes from the 32-robot Planar Cross scenario where neither {\arc} nor {\parc} achieve 100\% success.
However, we do see superlinear speedups in several scenarios with 100\% success rate by {\arc} and {\parc} with all tested worker allocations. 
While the batched subproblem resolution can change the set of conflicts discovered and subproblems attempted, the local OR-multi-start can further reduce wall-clock time by returning the first solution among multiple randomized repair attempts.
Consequently, the measured speedups are not necessarily bounded by the number of workers.
}

\subsubsection{Baseline Comparison}
\textcolor{black}{
In Mobile Cross and Planar Cross, where the conflicts are mostly independent, {\parc}-16 demonstrates the fastest runtimes and best scalability as the number of robots increases.
{\orparc}-4$\times$4 is the next-fastest and is more robust to unfavorable random seeds on the more complicated Planar Cross scenario where {\parc}-16 fails to solve some instances.
{\orarc}-16 also maintains a higher success rate than {\parc}-16 until the team size grows too large, further demonstrating the benefit of global OR-multi-start.
}

\textcolor{black}{
The baselines struggle with teams in the two cross scenarios with none solving either problem for 64 robots, and the composite baselines failing to complete any trials for the 16-robot Planar Cross scenario.
Even with parallelism, they often do not perform better than sequential {\arc}.
}

\textcolor{black}{
As expected, the {\arc} variants struggle on the Mobile Circle scenario. 
For the 4- and 8-robot scenarios, the EP-RRT variants report the fastest runtimes.
{\parc}-16 has the worst performance as there is little conflict independence to exploit its parallelism.
However, as the team size grows, the relative performance of {\orarc}-16 starts to surpass the composite methods. 
OR-EP-RRT-4$\times$4 and {\orparc}-4$\times$4 are nearly identical for 16 robots, 
and OR-RRT-C-16 consistently performs worse than EP-RRT-16. 
This demonstrates the benefit of OR-multi-start specifically for {\arc} variants.
OR-PP-ST-RRT scales best, maintaining 100\% success rate with subsecond solve times, indicating that the coupling natural in {\arc} methods is poorly matches to this scenario.
}

\textcolor{black}{
Finally, in the Panda Cage scenario, Figs.~\ref{fig:results}(j)-(l), we see the baselines struggle or fail to solve the problem as the team size grows while the {\arc} variants maintain high success rates (above $80\%$ even for 16 robots). 
The {\orarc} and {\orparc} variants maintain 100\% success rate on all problem sizes, and the hybrid {\orparc} variants demonstrate the greatest speedup in median runtime from sequential {\arc}. 
This supports the benefit of the hybrid parallel strategy to simultaneously improve robustness and runtime. 
}

\textcolor{black}{
We also observe the OR-multi-start {\arc} variants return the solutions with the lowest makespans (Table~\ref{tab:panda-results}).
This appears to be due to a correlation, for favorable seeds, between fewer conflicts and final makespan. 
Note that this is an observation only and a formal study of this is beyond the scope of this paper.
}

\subsection{Ablations}
\label{sec:ablations}

\textcolor{black}{
We run two ablations to study our design choices.
}

\subsubsection{Optimistically Independent Conflicts}
\label{sec:ablations-optimistic-independence}
\textcolor{black}{
We ran {\parc}-16 with strict conflict independence on the 8-robot Panda Cage scenario with the same settings as Section~\ref{sec:experimental-design}.
Table~\ref{tab:optimistic-independence} reports the comparative breakdown of the impact of optimistic versus strict independence on algorithm performance.
The average batch size increases slightly when including optimistically independent conflicts even in these smaller, often highly coupled problems.
The difference in conflict detection time is significant (more than 6$\times$ speed) and materially reduces the overall runtime (1.9$\times$ speedup on successful trials), justifying our inclusion of optimistically independent conflicts.
}

\begin{table}[t]
\centering
\caption{Impact of Optimistic Independence}
\label{tab:optimistic-independence}

\setlength{\tabcolsep}{4pt}
\renewcommand{\arraystretch}{1.12}

\begin{tabular*}{\columnwidth}{
  @{\extracolsep{\fill}}
  l
  l
  cc
  @{}
}
\toprule
Trial Set
& Metric
& Optimistic
& Strict \\
\midrule

\multirow{4}{*}{\makecell[l]{Successful\\Trials}}
& Count
& 47
& \textbf{48} \\

& Runtime (s)
& \textbf{6.25}
& 11.86 \\

& \makecell[l]{Conflict Detection (s)}
& \textbf{0.37}
& 2.32 \\

& \makecell[l]{Avg Batch Size}
& \textbf{1.51}
& 1.38 \\

& \makecell[l]{Avg First Batch Size}
& \textbf{2.43}
& 2.23 \\

\midrule

\multirow{3}{*}{\makecell[l]{All Trials (50)}}
& Runtime (s)
& \textbf{11.88}
& 15.34 \\

& \makecell[l]{Conflict Detection (s)}
& \textbf{0.36}
& 2.24 \\

& \makecell[l]{Avg Batch Size}
& \textbf{1.51}
& 1.39 \\

& \makecell[l]{Avg First Batch Size}
& \textbf{2.44}
& 2.26 \\

\bottomrule
\end{tabular*}

\vspace{2pt}
\begin{minipage}{\columnwidth}
\footnotesize
\raggedright
\textcolor{black}{
Results are reported for successful trials only and all trials (including timed-out trials--counted as 100 second runtimes) for the 8-robot Panda Cage scenario with 16 workers.
Runtime, conflict detection, batch size, and first batch size are all averages across the specified number of trials.
Runtime and conflict detection values are wall-clock times.}
\end{minipage}
\end{table}

\subsubsection{Synchronization Horizon}
\label{sec:ablation-sync-horizon}
\textcolor{black}{
We ran the first conflict detection call on the same 8-robot Panda scenario with varying synchronization horizon lengths $H$ for ten seeds on each of the five tasks while allowing optimistically independent conflicts (Table~\ref{tab:horizon-timing-results}). 
While there is little variation in the performance, there is a minimum conflict detection time at $H=200$ (which is the value used in the earlier Panda Cage experiments). 
}

\begin{table}[t]
\centering
\caption{Impact of Synchronization Horizon}
\label{tab:horizon-timing-results}

\setlength{\tabcolsep}{3pt}
\renewcommand{\arraystretch}{1.1}

\resizebox{\columnwidth}{!}{%
\begin{tabular}{@{}lrrrrrrr@{}}
\toprule
Horizon (timesteps)
& 50
& 100
& 200
& 400
& 800
& 1600
& 3200 \\
\midrule
Wall Time (ms)
& 10.15
& 9.93
& \textbf{9.85}
& 10.15
& 10.95
& 12.03
& 11.99 \\
Cumulative CPU Time (ms)
& 51.08
& \textbf{50.93}
& 55.14
& 64.08
& 88.25
& 120.66
& 120.00 \\
\bottomrule
\end{tabular}%
}

\vspace{2pt}
\begin{minipage}{\columnwidth}
\footnotesize
\raggedright
\textcolor{black}{
Average wall time and cumulative CPU time (in milliseconds) for the first parallel conflict detection call on the 8-robot Panda Cage scenario with 16 workers.
}
\end{minipage}
\end{table}

\section{Conclusion}
\label{sec:conclusion}
This paper presents Parallel ARC ({\parc}), exploiting the natural independence in {\arc}’s subproblem-based approach to {\mrmp} to distribute work between parallel workers.
The parallel processing improves planning time when the problem has sufficient independence in conflicts to distribute the work, but if this independence collapses, {\parc} can perform worse than sequential {\arc}.
Additionally, we study the impact of OR-multi-start strategies when applied to {\arc} and {\parc} and demonstrate improved success rates and often faster planning times and lower makespans, with hybrid {\orparc} performing best on the large, application-inspired multi-manipulator scenario.

\bibliographystyle{IEEEtran}
\bibliography{robotics.bib}

\end{document}